\pdfoutput=1

\documentclass[11pt]{article}

\usepackage[]{COLING2022}

\usepackage{times}
\usepackage{latexsym}

\usepackage[T1]{fontenc}

\usepackage[utf8]{inputenc}

\usepackage{microtype}

\usepackage{bm}

\usepackage{algorithm}
\usepackage{algorithmic}
\usepackage{enumerate}
\usepackage{CJKutf8}
\usepackage{url}
\usepackage{latexsym}
\usepackage{graphicx}
\usepackage{times}
\usepackage{latexsym}
\usepackage{amsmath}
\usepackage{array}
\usepackage{comment}
\usepackage{multirow}
\usepackage[english]{babel}
\usepackage{epstopdf}
\usepackage{times}
\usepackage{url}
\usepackage{latexsym}
\usepackage{graphicx}
\usepackage{amsmath}
\usepackage{amssymb}
\usepackage{amsmath}
\usepackage{mathrsfs}
\usepackage{booktabs}
\usepackage{mathtools}
\usepackage{pifont}
\usepackage[utf8]{inputenc}
\usepackage{algorithm}
\usepackage{algorithmic}
\usepackage{color}
\usepackage{ulem}
\usepackage{subfigure}
\usepackage{amssymb}
\usepackage{pifont}
\newcommand{\ourmethod}{\textsc{Paeg}}

\title{\ourmethod{}: Phrase-level Adversarial Example Generation for \\ Neural Machine Translation}

\author{
  Juncheng Wan\textsuperscript{\rm 1\thanks{\ Equal Contribution.}}, 
  Jian Yang\textsuperscript{\rm 2*}, 
  Shuming Ma\textsuperscript{\rm 3}, 
  Dongdong Zhang\textsuperscript{\rm 3}, 
  \\ {\bf Weinan Zhang}\textsuperscript{\rm 1}, 
  {\bf Yong Yu}\textsuperscript{\rm 1 \thanks{\ Corresponding author.}}, 
  {\bf Zhoujun Li}\textsuperscript{\rm 2} 
  \\
  \textsuperscript{\rm 1}Shanghai Jiao Tong University \\
  \textsuperscript{\rm 2}State Key Lab of Software Development Environment, Beihang University
  \\ 
  \textsuperscript{\rm 3}Microsoft Research Asia\\
  \{junchengwan,wnzhang,yyu\}@apex.sjtu.edu.cn; \\
  \{jiaya, lizj\}@buaa.edu.cn; \{shumma, haohua, dozhang\}@microsoft.com
}

\begin{document}
\maketitle
\begin{CJK*}{UTF8}{gbsn}
\begin{abstract}
While end-to-end neural machine translation (NMT) has achieved impressive progress, noisy input usually leads models to become fragile and unstable. Generating adversarial examples as the augmented data has been proved to be useful to alleviate this problem. Existing methods for adversarial example generation (AEG) are word-level or character-level, which ignore the ubiquitous phrase structure. In this paper, we propose a \textbf{P}hrase-level \textbf{A}dversarial \textbf{E}xample \textbf{G}eneration (\ourmethod{}) framework to enhance the robustness of the translation model. Our method further improves the gradient-based word-level AEG method by adopting a phrase-level substitution strategy. We verify our method on three benchmarks, including LDC Chinese-English, IWSLT14 German-English, and WMT14 English-German tasks. Experimental results demonstrate that our approach significantly improves translation performance and robustness to noise compared to previous strong baselines.
\end{abstract}

\section{Introduction}
Recently, neural machine translation (NMT) has effectively improved translation quality. NMT has shown state-of-the-art performance for many language pairs ~\citep{GNMT,human_parity,vaswani2017attention}. Various architectures~\citep{sutskever2014sequence, bahdanau2014neural,gehring2017convolutional,vaswani2017attention} bring many appealing properties. Most NMT systems heavily rely on high-quality parallel data and perform poorly in noisy input. With the noise rising in the source sentence, NMT tends to be more vulnerable~\citep{szegedy2013intriguing,DBLP:journals/corr/GoodfellowSS14}, due to the output prediction of the decoder easily intervened by the other words~\citep{DBLP:conf/acl/LiuTMCZ18}. A slight disturbance like a random permutation can damage the translation quality dramatically~\citep{DBLP:conf/iclr/BelinkovB18}. Even replacing a word with a synonym in the source input, the NMT model can be cheated and the target output can not be translated correctly.

\begin{CJK*}{UTF8}{gbsn}
\begin{table}[t]
\label{robust_example_1}
\begin{center}
\scalebox{1.00}{
\begin{tabular}{p{54pt}<{\centering}|p{140pt}}
\toprule 
Original Sentence   &  A cooked hot dog in a bun with ketchup and relish. \\ \midrule
Word-level AEG &  A cooked \textbf{\textit{\textcolor{blue}{warm}}} dog in a bun with ketchup and relish.  \\ \midrule
Phrase-level AEG &  A cooked \textbf{\textit{\textcolor{blue}{sausage rolls}}} in a bun with ketchup and relish.  \\ \bottomrule                    
\end{tabular}}
\caption{\label{example_1} An example of adversarial example generation (AEG). When the word ``hot'' is selected, word-level adversarial example generation method substitutes ``hot'' to ``warm''. The phrase-level method substitutes the whole phrase ``hot dog'' to ``sausage rolls''.}
\vspace{-10pt}
\end{center}
\end{table}

\end{CJK*}


To improve the robustness of the NMT model, previous works propose to construct the adversarial examples by manipulating hidden features or discrete text input. These adversarial examples are used as augmented data for the training of the NMT model. To attack hidden features, \citet{DBLP:conf/acl/LiuTMCZ18} added perturbations in the input at the feature level for adversarial stability training. To generate discrete adversarial input, \citet{ebrahimi2018adversarial} employed differentiable string-edit operations to rank adversarial changes.  \citet{DBLP:conf/iclr/BelinkovB18} and \citet{vaibhav2019improving} emulated naturally occurring errors in clean data as synthetic noise. \citet{cheng2019robust} proposed a gradient-based method to craft adversarial examples, considering the similarity between the gradient related to the translation loss of input and the embedding difference of words.


Previous methods of adversarial example generation (AEG) are limited at the low level, like word-level, not considering the relationship between different words within a phrase. There is one example in Table \ref{example_1}. The word-level AEG method selects a vulnerable position then substitutes the corresponding word, omitting that the substituted word is in a phrase. Sometimes, the examples of this improper substitution can even harm the translation model. 


Therefore, we propose a phrase-level adversarial example generation (\ourmethod{}) method, which improves a gradient-based word-level AEG method to phrase-level. Specifically, this method builds phrase-level candidates efficiently and substitutes phrases wholly with these candidates. We also propose to further improve this method with a bidirectional generation algorithm, as target-to-source adversarial pairs are a kind of slight perturbation of the original source-to-target translation. In practice, we generate adversarial examples after fixed intervals of NMT model updating (to convergence) and use them as new augmented data for the continual training of the model.

To verify the effectiveness of our method, we conduct experiments on three common benchmarks, i.e, LDC Chinese-English, IWSLT14 German-English, and WMT14 English-German. Experimental results demonstrate our method achieves significant improvements on translation quality and robustness to noisy inputs over the previous baselines including outstanding adversarial examples generation methods.

\section{Phrase-level Adversarial Example Generation}
In this section, we formulate the problem of adversarial example generation mathematically. First, our proposed method provides reliable candidates with the pre-trained model. Then, we use the gradient-based method to select vulnerable positions and substitutes at phrase level to generate adversarial examples. These examples are used as augmented data for the training of the NMT model. To further improve the performance, we extend our method to the bidirectional generation.

\subsection{Problem Formulation}
\label{formulation}
Let $A=\{(\textbf{x},\textbf{z}),\textbf{y}\}$ denotes the training data in NMT, where $(\textbf{x},\textbf{z})$ are the encoder input and the decoder input, $\textbf{y}$ the corresponding decoder output. To generate the corresponding adversarial examples $B=\{(\textbf{x}',\textbf{z}'),\textbf{y}\}$, where only the input is slightly different from $A$, we need to limit the adversarial input $(\textbf{x}',\textbf{z}')$ semantically close to the original data. 

Adversarial examples aim to cheat the model, making it predict wrong words. Therefore, given real output words $\textbf{y}$, we construct an input to make the model predict the incorrect word $\textbf{y}'(\textbf{y}'\neq \textbf{y})$. The process of adversarial example generation in NMT can be formulated as solving the following optimization problem:
\begin{gather}
	\{(\textbf{x}',\textbf{z}'):\mathop{\arg\max}_{(\textbf{x}',\textbf{z}')}P(\textbf{x}',\textbf{z}';\textbf{y},\theta),\nonumber \\ 
	dist((\textbf{x}',\textbf{z}'),(\textbf{x},\textbf{z}))<\epsilon \label{eq1}\}
\end{gather}
where $dist$ is a measure function of the input, such as the semantic distance of sentence embeddings or edit distance, $P(\textbf{x}',\textbf{z}';\textbf{y},\theta)$ is the maximal probability that the model predicts a wrong word $\textbf{y}'$ such that $\textbf{y}'\neq \textbf{y}$ when the model is fed with $(\textbf{x}',\textbf{z}')$, $\theta$ is the model parameters, and $\epsilon$ is a sufficiently small distance.

\begin{algorithm}[tb]
\caption{Phrase-level Adversarial Example}
\label{alg:algorithm1}
\textbf{Input}: $\{(\textbf{x}, \textbf{z}),\textbf{y}\}$ denotes input and output, $\theta_{l}^{s}$ and $\theta_{l}^{t}$ denote parameters of LMs, $\theta_{m}$ denotes parameters of the model, $\mathbb{D}$ denotes the phrase dictionary.\\
\textbf{Output}: phrase-level adversarial input: $(\textbf{x}', \textbf{z}')$.

\begin{algorithmic}[1] 
\STATE Compute $\{\textbf{g}_{x_i}\}_{i=1}^{|\textbf{x}|}$ with $\textbf{x}$, $\textbf{z}$, $\textbf{y}$ by Eq.(\ref{eq2}).
\STATE  $pos_x \longleftarrow$ positions of maximal $\{||\textbf{g}_{x_{i}}||_{2}\}_{i=1}^{|\textbf{x}|}$

\FOR{$i$ in $pos_x$}
    \STATE Get $cand(\textbf{x}_{ij})$ by $\mathbb{D}$ and $\theta_{l}^{s}$. 
    \STATE Substitute $\textbf{x}_{ij}$ to $\textbf{x}_{ij}'$ as Eq.(\ref{eq4}).
\ENDFOR
\STATE Compute $\{\textbf{g}_{z_i}\}_{i=1}^{|\textbf{z}|}$ with $\textbf{x}'$, $\textbf{z}$, $\textbf{y}$ by Eq.(\ref{eq2}).
\STATE Get attention matrix $\mathcal{M}$ by $\textbf{x}'$, $\textbf{z}$, $\textbf{y}$, and $\theta_{m}$.
\STATE Compute $\{P(j) \}_{j=1}^{|\textbf{y}|}$ with $\mathcal{M}$ by Eq.(\ref{eq3}).
\STATE $pos_z$ $\longleftarrow$ sampling by $\{P(j) \}_{j=1}^{|\textbf{y}|}$
\FOR{$i$ in $pos_z$}
    \STATE Get $cand(\textbf{z}_{ij})$ by $\mathbb{D}$, $\theta_{l}^{t}$ and $\theta_{m}$.
    \STATE Substitute $\textbf{z}_{ij}$ to $\textbf{z}_{ij}'$ as Eq.(\ref{eq4}).
\ENDFOR

\STATE \textbf{return} $(\textbf{x}', \textbf{z}')$
\end{algorithmic}
\end{algorithm}

\subsection{Phrase Candidates from PLM}
To guarantee the generated example $(\textbf{x}',\textbf{z}')$ is similar to the original example $(\textbf{x}, \textbf{z})$, two aspects are taken into account. One aspect is that the information in sentences should not change a lot. The other is to guarantee that words are similar. Therefore, high-quality candidates for words or phrases to be substituted should have similar semantic meanings to their original ones and be more fluent in the whole sentence.

To achieve this, one intuitive method is to select words with maximal prediction probability in the language model (LM), since LM predicts words based on the context. \citet{cheng2019robust} uses a bidirectional LM trained on the monolingual part of the parallel corpus. However, a high-quality LM often needs billions of monolingual data to train like BERT~\citep{DBLP:conf/naacl/DevlinCLT19}. It is unacceptable to spend much time and computational resources training reliable LMs. Therefore, we propose to utilize the knowledge of the pre-trained LM (PLM). In this paper, we use BERT as PLM.

In our paper, we use the notation $\textbf{x}_{ij}$ as the phrase from position $i$ to $j$ in sentence $\textbf{x}$, $cand(\textbf{x}_{ij})$ as the phrase candidates of phrase $\textbf{x}_{ij}$. When $i=j$, $\textbf{x}_{ij}$ indicates the word $x_i$ and $cand(\textbf{x}_{ij})$ indicates the word candidates $cand(x_i)$. Besides, we use $\mathbb{D}_n$ as the $n$-gram phrase dictionary and $\mathbb{D}$ the union set of all $\mathbb{D}_n$.

In the $i^{th}$ position of the source input, we construct $cand(x_i)$ by selecting the top $n_{s}$ tokens with maximal prediction probability in BERT when fed with $\textbf{x}$, where $x_i$ is masked. For the target input side, candidates consist of two parts. The first part is from BERT, which provides with $n_{t}^{l}$ candidates. The second part is from the trained NMT model, which provides with $n_{t}^{m}$ candidates. In this way, the candidate set $cand(z_j)$ of target input side consists of words fluent in the sentence and words conforming the translation of $\textbf{x}$. In this paper, we set $n_{s}^{l}=10$ and $n_{t}^{l}=n_{t}^{m}=5$.

Given a phrase $\textbf{x}_{ij}$, we construct phrase-level candidates $cand(\textbf{x}_{ij})$. We first build the set of all probable phrase candidates as the Cartesian product of all $cand(x_k)$ ($k=i, i+1,\dots,j$). Then, we screen out unreasonable phrase candidates by the phrase dictionary $\mathbb{D}$. Candidates not in this dictionary are discarded.

To obtain the phrase dictionary $\mathbb{D}$, we introduce two methods. The first one is to use the syntax parser to parse the sentence into a syntax tree. Then, the leaf nodes of an $n$-leaf subtree is an $n$-gram phrase. The phrase dictionary $\mathbb{D}$ is the union of these $n$-gram phrase dictionary $\mathbb{D}_n$. The second method is to utilize the existing phrase extraction tool directly. In this paper, we take both of these two methods. The syntax parser we used is {\tt nltk.parse}\footnote{\url{https://www.nltk.org}} and $n=2,3,4$. The phrase extraction tool we used is {\tt TextBlob}\footnote{\url{https://textblob.readthedocs.io/en/dev/}}. 

\subsection{Select Vulnerable Positions}
Instead of randomly selecting positions, we propose that the adversarial examples should select the most vulnerable positions in the sentence. Given a certain sentence, some NMT models may get worse translations when certain words or phrases are substituted.

Given that we train an NMT model with parameters $\theta_{m}$ and use negative log likelihood as the loss function with the input $\textbf{x}$, $\textbf{z}$ and the output $\textbf{y}$, we can get the gradient vector $\textbf{g}_{x_i}$ of token $x_i$ over the training loss:
\begin{equation}
	\textbf{g}_{x_i} = \nabla_{e(x_i)} -\log P(\textbf{y}|\textbf{x}, \textbf{z};\theta_{m})\label{eq2}
\end{equation}
where $e(x_i)$ is the embedding vector of token $x_i$. 


Previous methods randomly choose positions in the source input. Since different positions have different gradient norms $||\textbf{g}_{x_{i}}||_{2}$, if the gradient norm is large, the position is more unstable. Therefore, positions with large gradient norm are more vulnerable. For the source input, we select the top $\alpha_{s}|\textbf{x}|$ positions with maximal gradient norm, where $\alpha_{s}\in(0,1)$ is a ratio. \footnote{The experimental results of how to choose positions are discussed in Appendix \ref{vul_pos}.}

To construct the target input $\textbf{z}'$, we teach the model how to defend the attack from the source $\textbf{x}'$. It is a reason that we choose $n_{t}^{m}$ candidates from the NMT model on the target side. Selected target side positions should have the target counterpart of substituted source words in $\textbf{x}'$. For example, if we substitute the word ``drawing'' to ``eating'' in the source input ``Cezanne loved drawing apples .'' (``Cezanne malt gerne {\"a}pfel .'' in German), then we need to find the position of the corresponding translation ``drawing'' (``malt'') and substitute it to an English word related to ``eating'', such as ``isst''. 


This process is the inverse process of attention in NMT. Following \cite{cheng2019robust}, we sample $\alpha_{t}|\textbf{y}|$ ($\alpha_{t}\in(0,1)$) relevant words influenced by the perturbed words in the source input $\textbf{x}'$ as by sampling function $P(\cdot)$:
\begin{eqnarray}
	P(j) = \frac{\sum_i \mathcal{M}_{ij}\delta_{x_i\ne x_i'}}{\sum_k\sum_i\mathcal{M}_{ik}\delta_{x_i\neq x_i'}}, j=1,\dots,|\textbf{y}|\label{eq3}
\end{eqnarray}where $\mathcal{M}_{ij}$ is the value of attention matrix between token $x_i$ and token $y_j$ from NMT model, $\delta_{x_i\neq x_i'}$ is 1 if $x_i \neq x_i'$ and 0 otherwise.

\subsection{Phrase-level Substitution}
Since words in the same phrase have a close relationship, we substitute words at phrase level in the adversarial example generation. There are two aspects to consider. First, since synonymy phrases sharing the same meaning may have variant lengths, the feature representation of a phrase should be irrelevant to the length of the phrase. Besides, we need to choose the phrase from the candidate set that disturbs the model the most. 

For the first consideration, we simply extract phrase-level features by averaging the word embeddings. For the second aspect, we adopt the gradient-based approach in~\citep{cheng2019robust}. To represent the whole gradient of the phrase, we also average all the gradients of words. Other feature engineering methods like max-pooling, concatenation, and element-wise product are also viable.

Formally, for the substitution of phrase $\textbf{x}_{ij}$, the greedy approach based on the gradient is:
\begin{eqnarray}
\textbf{x}_{ij}' = \mathop{\arg\max}_{\textbf{c}\in cand(\textbf{x}_{ij})}sim(f^e(\textbf{c})-f^e(\textbf{x}_{ij}),f^g(\textbf{c}))\label{eq4}
\end{eqnarray}
where $sim$ is the similarity function, $f^e$ is the feature representation of the phrase, $f^g$ is the feature representation of the gradient of the phrase, and $cand(\textbf{x}_{ij})$ the phrase candidates of $\textbf{x}_{ij}$. In this paper, we use the ``average'' function for $f^e$ and $f^g$ \footnote{The reason for using ``average'' function is explained in Appendix \ref{phrase_embed}.} and cosine similarity as the similarity function.

Our phrase-level adversarial example generation process is shown in Algorithm \ref{alg:algorithm1}. During the training of the NMT model, we generate adversarial examples periodically as augmented data. Note that we do not need training LMs for the source and target languages.

\begin{algorithm}[tb]
\caption{Bidirectional Generation}
\label{alg:algorithm2}
\textbf{Input}: $\{(\textbf{x}, \textbf{z}_l),\textbf{y}\}$ denotes source-to-target input and output. $\{(\textbf{y}, \textbf{z}_r),\textbf{x}\}$ denotes target-to-source input and output. \textbf{Gen} is the adversarial examples generator. \\
\textbf{Output}: augmented source-to-target data $D_l$ and target-to-source data $D_r$.

\begin{algorithmic}[1] 
\STATE Compute $\{\textbf{g}_{x_i}\}_{i=1}^{|\textbf{x}|}$ with $\textbf{x}$, $\textbf{z}$, $\textbf{y}$ by Eq.(\ref{eq2}).
\STATE $D_l \longleftarrow \{(\textbf{x}, \textbf{z}_l),\textbf{y}\}$, $D_r \longleftarrow \{(\textbf{y}, \textbf{z}_r),\textbf{x}\}$.
\STATE  $\textbf{x}', \textbf{z}_{l}' \longleftarrow \text{Gen}(\textbf{x}, \textbf{z}_l)$,  $\textbf{y}', \textbf{z}_{r}' \longleftarrow \text{Gen}(\textbf{y}, \textbf{z}_r)$
\STATE  Add $\{(\textbf{x}', \textbf{z}_l'),\textbf{y}\} \bigcup \{(\textbf{z}_r', \textbf{y}'),\textbf{y}\} $ to $D_l$ and \\ add $\{(\textbf{y}', \textbf{z}_r'),\textbf{x}\} \bigcup \{(\textbf{z}_l', \textbf{x}'),\textbf{x}\}$ to $D_r$.
\STATE \textbf{return} $(D_l, D_r)$
\end{algorithmic}
\end{algorithm}

\subsection{Bidirectional Generation}
In practice, reversed adversarial examples from target-to-source translation can also be used as augmented data for the source-to-target translation. Therefore, we introduce a bidirectional generation method to boost our phrase-level adversarial example generation method.

Our bidirectional generation has two translation directions, source-to-target, and target-to-source. We use a universal encoder and decoder for these two directions as~\citep{johnson2017google}. From the original data, we generate the adversarial examples for two directions. In each iteration, the adversarial examples are reversed and added to the dataset. The model is trained on the augmented dataset.

Formally, we notate $(\textbf{x}, \textbf{z}_{l}, \textbf{y})$ as the encoder input, decoder input and decoder output for source-to-target translation, $(\textbf{y}$, $\textbf{z}_{r}$, $\textbf{x})$ as the encoder input, decoder input and decoder output for target-to-source translation. After generating the adversarial examples, we get $(\textbf{x}', \textbf{z}_{l}', \textbf{y})$ and $(\textbf{y}', \textbf{z}_{r}'$, $\textbf{x})$. Then, the adversarial examples input are reversed and added to the training data of the other direction. For source-to-target training, we have three pairs of data $(\textbf{x},\textbf{z},\textbf{y}), (\textbf{x}',\textbf{z}_{l}',\textbf{y}), (\textbf{z}_{r}',\textbf{y}',\textbf{y})$. They are respectively the original training data, the adversarial examples and the reversed adversarial examples from the other direction. 

The phrase-level adversarial example generation of these two directions help mutually during the training. Our bidirectional generation algorithm is shown in Algorithm \ref{alg:algorithm2}. It is worth noting that, in general, the training time of \ourmethod{} is not as much as double of \ourmethod{} without bidirectional generation, as the data from bidirectional generation has a similar distribution of the original adversarial samples.\footnote{There are more discussions about time consumption of bidirectional generation in Appendix \ref{train_time}.} 


\begin{table*}[!t]
\begin{center}
{
\scalebox{0.85}{
\begin{tabular}{l|c|ccccc|c}
\toprule
Method                            & MT06 & MT02 & MT03 & MT05 & MT08 & MT12 & Avg. \\ \midrule
Transformer~\citep{vaswani2017attention}       & 43.52 & 43.17 & 44.06 & 44.45 & 36.27 & 35.07 & 41.09  \\ 
Multilingual NMT~\citep{johnson2017google}        & 43.54 & 43.46 & 44.63 & 44.40 & 36.13 & 35.00 & 41.19  \\ \midrule
Word Dropout~\citep{WordDropout}\dag  & 43.96 & 44.02 & 44.55 & 44.70 & 36.49 & 35.33 & 41.51  \\
SwitchOut~\citep{DBLP:conf/emnlp/WangPDN18}\dag & 43.83 & 44.36 & 45.02 & 44.85 & 36.53 & 35.45 & 41.67  \\
AdvGen~\citep{cheng2019robust}\dag        & 44.74 & 45.12 & 46.49 & 45.95 & 37.29 & 36.02 & 42.60  \\ \midrule
\textbf{\ourmethod{} (this work)}\dag   & \textbf{45.49} & \textbf{45.76} & \textbf{47.58} & \textbf{46.83} & \textbf{38.18} & \textbf{36.91} & \textbf{43.46}  \\ \bottomrule
\end{tabular}
}}
\caption{\label{LDC-zh2en} Case-insensitive BLEU-4 scores (\%) on LDC Zh$\to$En task. Our method is compared with other baselines and \textit{Transformer\_base} model. Methods with ``\dag'' use adversarial examples for training. 
}
\vspace{-10pt}
\end{center}
\end{table*}

\begin{table}[!t]
\begin{center}
\scalebox{0.85}{
\begin{tabular}{l|c}
\toprule
Method                            & \multicolumn{1}{c}{BLEU}       \\ \midrule
Transformer~\citep{vaswani2017attention}      & 34.20       \\ 
Multilingual NMT~\citep{johnson2017google}   & 34.13       \\ 
NT$^2$MT~\citep{DBLP:journals/corr/abs-1811-02172} & 31.75       \\ 
LightConv~\citep{LightConvAndDynamicConv}      & 34.80  \\
DynamicConv~\citep{LightConvAndDynamicConv}    & 35.20  \\ \midrule
Word Dropout~\citep{WordDropout}\dag  & 34.72  \\
SwitchOut~\citep{DBLP:conf/emnlp/WangPDN18}\dag & 34.83  \\
AdvGen~\citep{cheng2019robust}\dag     & 35.25 \\  \midrule
\textbf{\ourmethod{} (this work)}\dag    & \textbf{35.65} \\ \bottomrule
\end{tabular}
}
\caption{\label{IWSLT-de2en} Case-insensitive BLEU-4 scores (\%) on IWSLT14 De$\to$En task. Our method is compared with other baselines and \textit{Transformer\_small} model. Methods with ``\dag'' use adversarial examples for training.
}
\vspace{-10pt}
\end{center}
\end{table}

\begin{table}[!t]
\begin{center}
\scalebox{0.85}{
\begin{tabular}{l|c}
\toprule
Method                            & \multicolumn{1}{c}{BLEU}       \\ \midrule
Transformer~\citep{vaswani2017attention}      &   28.40    \\ 
Multilingual NMT~\citep{johnson2017google}   &     29.11   \\
RNMT+~\citep{RNMT+} &    28.49    \\
LightConv~\citep{LightConvAndDynamicConv}      & 28.90 \\
DynamicConv~\citep{LightConvAndDynamicConv}    & 29.70 \\ \midrule
Word Dropout~\citep{WordDropout}\dag  &  29.30 \\
SwitchOut~\citep{DBLP:conf/emnlp/WangPDN18}\dag  & 29.40   \\
AdvGen~\citep{cheng2019robust}\dag    &  30.01 \\  \midrule
\textbf{\ourmethod{} (this work)}\dag   & \textbf{30.49} \\ \bottomrule
\end{tabular}
}
\caption{\label{WMT-en2de} Case-insensitive BLEU-4 scores (\%) on WMT14 En$\to$De task. Our method is compared with other baselines and \textit{Transformer\_big} model. Methods with ``\dag'' use adversarial examples for training.  
}
\vspace{-20pt}
\end{center}
\end{table}

\begin{table*}[!t]
\begin{center}{
\scalebox{0.85}{
\begin{tabular}{l|c|ccccc|c}
\toprule
Method                         & MT06  & MT02 & MT03  & MT05  & MT08  & MT12 & Avg. \\ \midrule
\ourmethod{}                           & 45.49 & 45.76 & 47.58 & 46.83 & 38.18 & 36.91 & 43.46  \\
w/o bidirectional generation          & 45.52 & 45.53 & 46.96 & 46.72 & 38.10 & 36.85 & 43.24 \\ 
w/o phrase-level substitution    & 44.03 & 44.02 & 45.63 & 45.35 & 37.21 & 35.51 & 41.96 \\ 
w/o candidates from BERT            & 43.52 & 43.17 & 44.06 & 44.45 & 36.27 & 35.07 & 41.09 \\
\bottomrule
\end{tabular}
}}
\caption{\label{abla} Experiments on LDC Zh$\to$En dataset to analyze the effect of different components of \ourmethod{}. We removed three components of \ourmethod{} step by step. The results show that phrase-level substitution is the most effective part.}
\vspace{-20pt}
\end{center}
\end{table*}

\section{Experiments}
We evaluate our method on three datasets, LDC Chinese-English and IWSLT14 German-English translation datasets. Then, we compare our method with baselines. At last, we do a detailed analysis of the different components of our method.

Limited by the number of pages, we have included the description of three datasets and the training details in the Appendix \ref{dataset} and Appendix \ref{train_details} respectively.

\subsection{Comparisons to Baseline Methods}
We compare our method with NMT models without adversarial examples (Non-adv NMT) and using adversarial examples (Adv NMT). 
Our method gets significant translation improvement by statistical significance testing ($p<0.05$) compared to relevant baselines.

\paragraph{Non-adv NMT} \textbf{Multilingual NMT} \cite{johnson2017google} is implemented with the Transformer model as the universal encoder and decoder. \textbf{NT$^2$MT} \cite{DBLP:journals/corr/abs-1811-02172} uses a phrase attention mechanism with backbone model LSTM. We report the maximal result with out-of-domain dictionaries in the paper.
\textbf{RNMT+} \cite{RNMT+} is an enhanced version of RNN-based NMT model.
\textbf{LightConv} \cite{LightConvAndDynamicConv} uses a lightweight convolution performing competitively to the Transformer. 
\textbf{DynamicConv} \cite{LightConvAndDynamicConv} leverages a dynamic convolution predicting separate convolution kernels. 

\paragraph{Adv NMT} \textbf{Word Dropout} \cite{WordDropout} drops words randomly. We implement it on the token level, as recommended by the paper. \textbf{SwitchOut} \cite{DBLP:conf/emnlp/WangPDN18} randomly replaces words in both the source and target sentence with words from the vocabulary. We implement the hamming distance sampling method in the paper. \textbf{AdvGen} \cite{cheng2019robust} is an adversarial example generation method at the word-level. This method uses doubly adversarial input. We implement this method with the Transformer backbone, $\alpha_{s}=25\%, \alpha_{t}=50\%$ for LDC Chinese-English task, and $\alpha_{s}=20\%, \alpha_{t}=20\%$ for IWSLT14 German-English and WMT14 English-German.

Table \ref{LDC-zh2en} demonstrates the comparisons between our method with the above five baseline methods on LDC Chinese-English translation task. First, we compare our method with the Transformer. On average, \ourmethod{} can improve +2.37 BLEU points significantly. Then, we compare our method with methods of training with adversarial examples. On average, adversarial example generation methods (AdvGen and \ourmethod{}) utilizing the training information of the model greatly surpass the other methods (Word Dropout and SwitchOut). The reason is that the former approach is better at attacking vulnerable parts of the NMT model. Compared with the state-of-the-art AEG method AdvGen, \ourmethod{} gets an improvement of +0.86 BLEU points.

In Table \ref{IWSLT-de2en}, we compare our method with the above eight baseline methods on the IWSLT14 German-English translation task. Compared with the backbone model Transformer, \ourmethod{} gets the gain of +1.45 BLEU points. Compared with methods built on top of Transformer, NT$^2$MT~\citep{DBLP:journals/corr/abs-1811-02172} with out-of-domain dictionaries suffers from a worse backbone model (LSTM). Multilingual NMT~\citep{johnson2017google} has a similar performance to the Transformer model. Compared with the other methods of training with adversarial examples, \ourmethod{} has the best performance. \ourmethod{} gets +0.8$\sim$0.9 BLEU points improvement compared with AEG methods which do not leverage the training information of the model.

The comparisons on the WMT14 English-German task are in Table \ref{WMT-en2de}. Compared with \textit{Transformer\_big} model, \ourmethod{} has a notable gain of +2.09 BLEU points. \ourmethod{} consistently outperforms all three baselines training with adversarial examples, having around +0.5$\sim$1.0 BLEU points improvement in this commonly used dataset.


\subsection{Ablation Studies}
Our proposed method \ourmethod{} is mainly affected by three components, the use of the pre-trained model, the phrase-level substitution, and the bidirectional adversarial example generation. We analyze the different components of \ourmethod{} by ablation studies.

\paragraph{Effect of Phrase-level Substitution}
We use the phrase-level substitution and there is +1.28 BLEU points improvement in Table \ref{abla}, which is significant. Substituting words randomly from the top 10 word-level candidates can not guarantee consistency between words. What is worse is that random substitution may destroy the phrase structure and semantic consistency in the sentence. 

For common languages, such as Chinese, English, and German, the ratio of phrases is non-negligible. Substituting at the phrase level does make the adversarial input more fluent and thus more closely approximates the real-world data. In this way, our method can teach the model to defend against the attack on the target side better.

\paragraph{Effect of PLM}
To find out the impact of the pre-trained model, we use BERT to generate pseudo data. In Table \ref{abla}, with the use of the pre-trained model BERT, the Transformer model has +0.87 BLEU points improvement. This proves that BERT provides more reliable candidates by pre-training on amounts of data. Compared with the LMs trained on millions of monolingual data, BERT can significantly leverage the contextual information to make the candidates appear fluent in the sentence.

\paragraph{Effect of Bidirectional Generation}
In Table \ref{abla}, we add the bidirectional generation method to \ourmethod{} and there is +0.22 BLEU points improvement. This shows that the bidirectional generation has slight improvements. Considering that PAEG (without bidirectional generation) itself achieves a high BLEU score, the further improvement of bidirectional generation cannot be ignored.  


\subsection{Robustness to Noisy Inputs}\label{robustness_to_noisy_inputs}

To compare the robustness of different NMT models, we conduct three groups of experiments to simulate machine translation scenarios with noisy inputs by word replacement and switch. All experiments are conducted in the WMT14 English-German test set. Our method is compared with the representative word-level augmentation method AdvGen with the \textit{Transformer\_big} backbone. 

\begin{figure}[t]
    \centering
    \subfigure[]{
    \includegraphics[width=0.46\columnwidth]{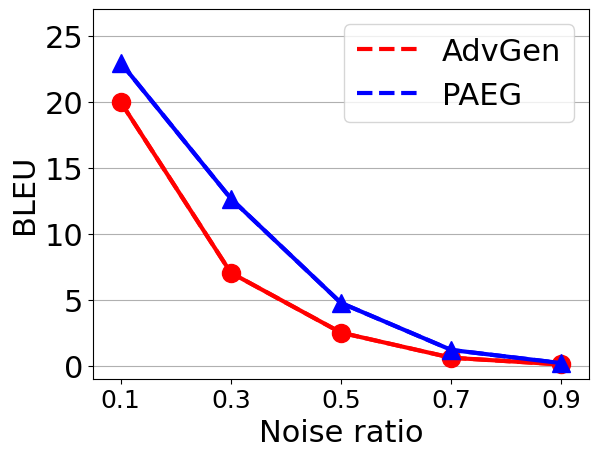}
    \label{rrn}
    }
    \subfigure[]{
    \includegraphics[width=0.46\columnwidth]{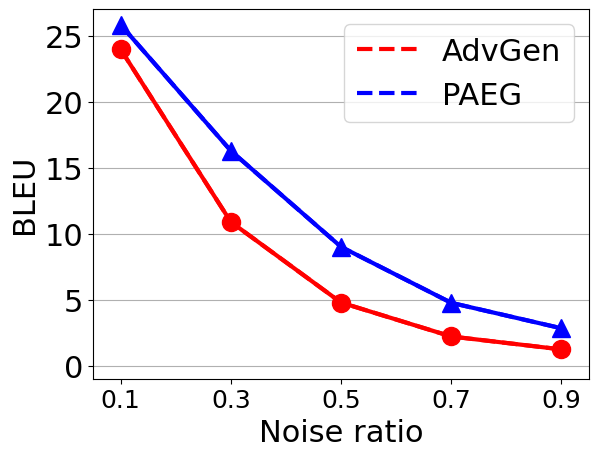}
    \label{rsn}
    }
    \caption{Results on (a) random replacement noise and (b) random switch noise.} 
    \vspace{-10pt}
\end{figure}



\paragraph{Random Word Replacement/Switch Noise} 
We first simulate the random replacement and switch noise, where a specific proportion ($\gamma$) of \textit{positions} of the source sentence are selected uniformly and replace with random words in the source vocabulary (also uniformly). Such phenomenon is common in real-world scenarios, like onomatopoeia in speech recognition. We set $\gamma\in \{ 0.1, 0.3, 0.5, 0.7, 0.9\}$ to indicate the level of noise and test the sensitivity of NMT models in Figure \ref{rrn} and Figure \ref{rsn}. 

The analytic results show that our method \ourmethod{} improves the robustness of NMT models more than AdvGen, both to random replacement noise and switch noise. When the ratio of noise increases, the BLEU improvement gets consistently larger, which proves the effectiveness of \ourmethod{}. When the ratio of noise is high (0.7$\sim$0.9), both methods degenerate into random translation machines. It can be attributed that excessive random noise impairs the source-side encoding.

\begin{figure}[t]
    \centering
    \subfigure[]{
    \includegraphics[width=0.46\columnwidth]{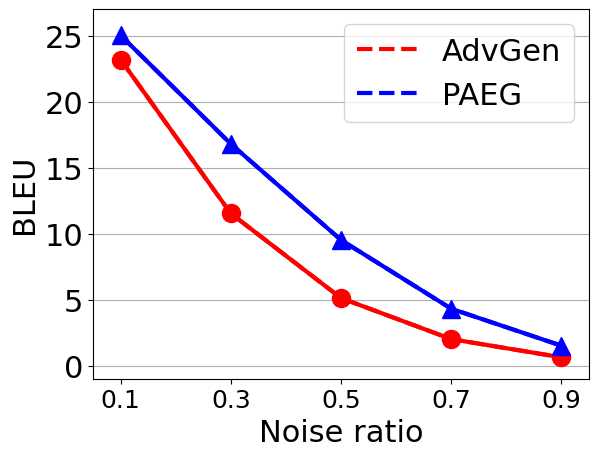}
    \label{msrn}
    }
    \subfigure[]{
    \includegraphics[width=0.46\columnwidth]{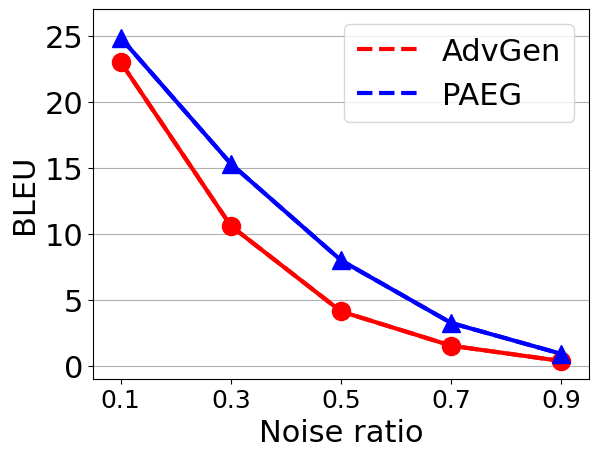}
    \label{lsrn}
    }
    \caption{Results on (a) word-level most similar synonym noise and (b) word-level least similar synonym noise.} 
    \vspace{-5pt}
\end{figure}



\paragraph{Word-level Synonym Noise} 
Another common noise in the translation system is synonym substitution, where the translation system is required to translate sentences consistently with subtle synonym difference. We first simulate this scenario with moderate word replacements. The selection of noisy positions is the same as random word replacement. Each to-be-replaced word matches the top 5 similar words by word similarity as the candidate set.\footnote{We use word embedding cosine similarity by pre-trained word embeddings GloVe (100 dimension) from flairNLP.} We add the most/least similar synonym noise by selecting the most/least similar word in the candidate set as the replacement. The noise ratio $\gamma\in \{ 0.1, 0.3, 0.5, 0.7, 0.9\}$ and sensitivity results are in Figure \ref{msrn} and Figure \ref{lsrn}. 

The noise-BLEU curves have almost the same trend as random word replacement, which again proves the superior robustness of \ourmethod{} over AdvGen at word-level synonym noises. This is understandable because \ourmethod{} is inclusion of word-level alternatives.

\begin{figure}[t]
    \centering
    \subfigure[]{
    \includegraphics[width=0.46\columnwidth]{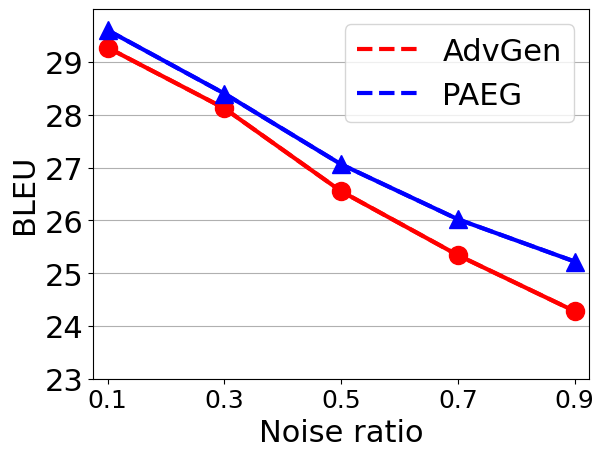}
    \label{pmsrn}
    }
    \subfigure[]{
    \includegraphics[width=0.46\columnwidth]{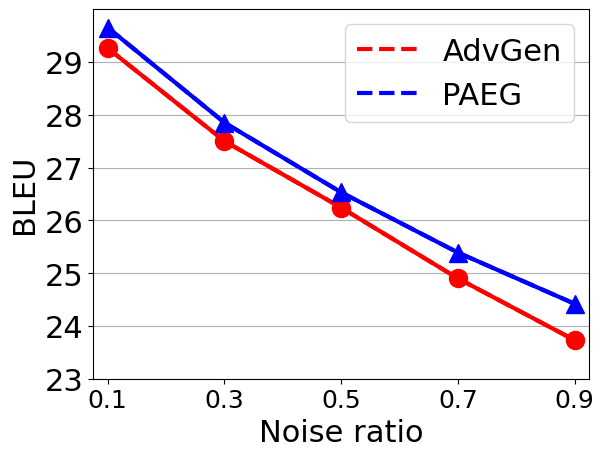}
    \label{plsrn}
    }
    \caption{Results on (a) phrase-level most similar synonym noise and (b) phrase-level least similar synonym noise.} 
    \vspace{-5pt}
\end{figure}



\paragraph{Phrase-level Synonym Noise} 
In addition, we would like to verify how robust our method is to phrase-level synonym noises, where phrase structures are destroyed by word-level synonyms replacement, such as the case in Table \ref{example_1}. For this purpose, we select $\epsilon$ ratio of \textit{phrases} uniformly and replace them with similar words in the source language vocabulary. The noise ratio $\epsilon\in \{ 0.1, 0.3, 0.5, 0.7, 0.9\}$. Figure \ref{pmsrn} and Figure \ref{plsrn} show that with the increase of phrase-level noise, \ourmethod{} gets more BLEU improvement both in most and least similar synonym noise settings. Our method is more resistant to the destruction of phrase structures, which is proved again in the following section.

\subsection{Analysis of Phrase-level Substitution}

\begin{table}[!t]
\begin{center}{
\resizebox{0.75\columnwidth}{!}{
\begin{tabular}{c|cc}
\toprule
Word Type      & Word-level & Phrase-level  \\ \midrule
NP (\%)      & 18.69    &  \ 20.11 (\textbf{+1.42})   \\
VP (\%)      &  \; 7.56    & \; 7.48 (-0.08)  \\
PP (\%)      & 16.43    & \ 17.53 (+1.10)\;  \\
ADJP (\%)    & \; 1.76    & \; 1.98 (+0.22) \\ \bottomrule
\end{tabular}
}}
\caption{\label{dif_phra} Experiments on the LDC Zh$\to$En task to compare the phrase-level and word-level AEG methods by the ratio of noun/verb/prepositional/adjective phrases (NP/VP/PP/ADJP), in the hypothesis.}
\vspace{-10pt}
\end{center}
\end{table}

\begin{table}[!t]
\begin{center}{
\resizebox{0.8\columnwidth}{!}{
\begin{tabular}{l|cc}
\toprule
$N$-gram   & Word-level & Phrase-level  \\ \midrule
1-gram BLEU  & 79.56    & 79.78 (+0.22)   \\
2-gram BLEU   & 52.87    & 53.59 (+0.72) \\
3-gram BLEU   & 34.49    & 35.65 (\textbf{+1.16}) \\
4-gram BLEU   & 24.22    & 25.03 (+0.81)  \\ \bottomrule
\end{tabular}
}}
\caption{\label{bleu4} Experiments on the LDC Zh$\to$En to compare the phrase-level and word-level AEG method in $n$-gram BLEU scores. Phrase-level method improves $n$-gram ($n>1$) BLEU scores more.}
\vspace{-20pt}
\end{center}
\end{table}



\begin{table*}[!t]
\begin{center}{
\scalebox{0.82}{
\begin{tabular}{l|l}
\toprule
\toprule
Original   & SRC: 对正在实施的(家庭/family)(暴力/violence)，(受害人/victim)可以(请求/ask)公安(机关/organ)救助。  \\
           & TGT: With regard to ongoing family violence, the victim may ask the public security organ for help. \\ \midrule
AdvGen     & SRC: 对正在实施的(家庭/family)(\uline{迫害}/abuse)，(受害人/victim)可以(\uline{要求}/require)公安(机关/organ)救助。   \\
           & TGT: With regard to ongoing family \textbf{\uline{abuse}}, the \textbf{\uline{victim}} may \textbf{\uline{require}} the public security organ for help. \\ \midrule
\ourmethod{}       & SRC: 对正在实施的(家庭/family)(\uline{虐待}/abuse)，(受害人/victim)可以(请求/ask)公安(\uline{警察}/police)救助。 \\
           & TGT: With regard to ongoing \textbf{\uwave{domestic abuse}}, the \textbf{\uline{sufferer}} may ask the public security \textbf{\uline{police}} for help. \\ \bottomrule
\bottomrule
\end{tabular}
}
}
\caption{\label{case_study} Comparison of our \ourmethod{} method and AdvGen method on the LDC Zh$\to$En dataset. Tokens with underline are substituted by the model as a word. Tokens with wave lines are substituted by the model as a phrase entirely. Chinese tokens and their English counterparts are in brackets (Chinese/English).}
\vspace{-20pt}
\end{center}
\end{table*}
Phrase-level substitution shows remarkable improvement of the BLEU scores on average. In this subsection, we analyze the translation details and discuss the reason for such an improvement. 

\paragraph{Phrase Translation}
First, we make the statistics of the ratio of phrases $\eta$ of the (generated) hypothesis in LDC Chinese-English translation in Table \ref{dif_phra}. In a text $\textbf{x}$, the ratio of phrases $\eta$ is defined as the sum of the phrase lengths in $\textbf{x}$ divided by the text length $|\textbf{x}|$. For the word-level AEG method, the $\eta$ of noun phrases (NP) is $18.69\%$ on average. While for the phrase-level method, the ratio of NP is remarkably $20.11\%(+1.42\%)$. Besides, the $\eta$ of prepositional phrases (PP) also increases $1.1\%$ by phrase-level substitution. 

These results show that the NMT model trained on \ourmethod{} considers more about phrases, especially NP and PP. Phrase-level substitution prevents the damage to the structure of phrases, guarantee the normal ratio of phrases in the augmented dataset, and thus teaches the decoder to generate phrases.

\paragraph{$N$-gram Accuracy}
Besides, we analyze the improvements for different $n$-gram BLEU scores in Table \ref{bleu4}. \ourmethod{} improve the 3-gram BLEU greatly (+1.16 points) over the word-level method. 2-gram and 3-gram BLEU also get moderate improvements (+0.7$\sim$0.8 points), much greater than 1-gram BLEU. These results verify that, using phrase-level strategy, longer grams can be translated more accurately (to match the phrases in the references).   

\paragraph{Case Study}
In Table \ref{case_study}, there is a case of the example generating process from the LDC dataset. On the target side, AdvGen substituted ``violence'' to ``abuse''. \ourmethod{} selected the $6$-th position of the target sentence and substituted ``family violence'' to ``domestic abuse'' entirely. Though ``family abuse'' does not violate the original meaning, the substitution ``domestic abuse'' is more reasonable.

\section{Related Work}

Adversarial training for neural networks has been studied recently~\citep{szegedy2013intriguing, DBLP:journals/corr/GoodfellowSS14}. Similar ideas are applied into natural language processing~\citep{lamb2016professor,li2017adversarial,yang2018improving,DBLP:conf/acl/LiuTMCZ18,cheng2019robust,adv_aug,DBLP:conf/acl/NamyslBK20,DBLP:conf/acl/CroceCB20,DBLP:conf/acl/WangHZHXYX20,DBLP:conf/acl/ZangQYLZLS20,DBLP:conf/acl/DingLXZXWZ20}. Specifically, adversarial example generation \cite{fadaee2017data,ebrahimi2018adversarial,wang2018switchout,adv_aug,DBLP:conf/acl/ZouHXDC20,DBLP:conf/acl/ZhengZZHCH20,DBLP:conf/acl/HideyCAVKDM20} is proved to be useful to train a robust NMT system. Recently, \citet{cheng2019robust} adopted a gradient-based method to craft adversarial examples at word level, using the adversarial source input to attack while the target input to defend the model.


Our bidirectional generation method is similar to multilingual NMT training. Multilingual NMT models \cite{DBLP:conf/acl/DongWHYW15,DBLP:journals/corr/LuongLSVK15,johnson2017google,DBLP:conf/acl/WangTN20,DBLP:conf/acl/ZhangWTS20,DBLP:conf/acl/ZhuYCL20,DBLP:conf/acl/SiddhantBCFCKAW20} are trained over multiple language pairs with parameter sharing, such as using the same encoder/decoder for different source/target languages \cite{johnson2017google}, using one encoder and separate decoders to translate one language to multiple languages \cite{DBLP:conf/acl/DongWHYW15}, and sharing an attention mechanism \cite{DBLP:conf/naacl/FiratCB16} across multiple language pairs. In this work, we use the adversarial examples generated from the other direction to improve the robustness of the original translation direction.




\section{Conclusion}
In this work, we propose a phrase-level adversarial example generation method. Our goal is to improve the fluency of the adversarial examples. We improve a gradient-based word-level method with phrase-level candidate construction, overall substitution strategy, and bidirectional generation. We verify our method on Chinese-English, German-English, and English-German corpus, and the results show that \ourmethod{} can improve both translation quality and robustness to noisy inputs significantly. 




\normalem
\bibliography{anthology,custom}
\bibliographystyle{acl_natbib}

\newpage
\appendix


\section{Details of \ourmethod{}}
\subsection{Vulnerable Positions}\label{vul_pos}
In this work, there is an assumption that substituting words in vulnerable positions (positions with greater gradient norm) is more likely to add perturbation to model training. In our experiments, we have tried sampling the positions of source phrases randomly and found that vulnerable positions is better (+0.2$\sim$0.3 BLEU points).

\subsection{Phrase Embedding}\label{phrase_embed}
In the experiments, ``max-pooling'' has been explored to get the phrase embedding/gradients from word embedding/gradients, and it has a similar result as the ``average'' operation, within 0.2 BLEU points. In the implementation, ``max-pooling'' is slower than the ``average'' (using PyTorch 1.7), therefore we choose ``average'' for convenience.


\section{Dataset}\label{dataset}
\paragraph{LDC Chinese-English Task}
This is a dataset of 1.2M training sequence pairs. The LDC numbers are 2002E17, 2002E18, 2004T08,  2005T10, 2005T34, 2006E17, 2006T06, and 2008T18\footnote{\url{https://catalog.ldc.upenn.edu/byproject}}. We choose the NIST 2006 as the validation set, which has 1664 sentences, and the NIST 2002, NIST 2003, NIST 2005, NIST 2008, NIST 2012 as the test sets, which contain 877, 919, 1082, 1357, 2190 sentences respectively.

\paragraph{IWSLT14 German-English Task} 
This dataset comes from translated TED talks. This dataset contains roughly 160K pairs as the training set, 7K pairs as the validation set, and 7K pairs as the test set, respectively. We take the IWSLT14 test set as the test set.

\paragraph{WMT14 English-German Task} 
The training data has 4.5M sentence pairs. We use the newstest2013 as the valid set and the newstest2014 as the test set.

\section{Training Details}\label{train_details}
Our backbone model is the Transformer model \citep{vaswani2017attention}. The NMT model consists of a Transformer encoder and a Transformer decoder. The pre-trained LM is BERT-based\footnote{\url{https://github.com/huggingface/transformers}}. We use {\tt nltk.parse} to build the syntax tree and extract the phrases of length $2,3,4$. Besides, we use {\tt TextBlob} to extract the noun phradses and merge other phrases (from {\tt nltk.parse}) to build the phrase dictionary.


\paragraph{LDC Chinese-English Translation}
We use our in-house Chinese word-breaker toolkit to segment Chinese data. We use byte pair encoding (BPE) to encode sentences with a shared token vocabulary of 51K sub-word tokens. The size of the phrase vocabulary is 1.2M for Chinese and 0.9M for English. We limit the maximum sentence length up to 256 words. We apply Adam~\citep{DBLP:journals/corr/KingmaB14} with $\beta_1=0.9$ and $\beta_2=0.98$ to train models for 80 epochs and select the best model parameters according to the model performance on the valid set. We use \textit{Transformer\_base} setting: embedding size as 512, feed-forward network (FFN) size as 2048, attention heads as 8, learning rate as 0.1, batch size as 6144, and dropout rate as 0.1. We use the warm-up strategy with 4000 warm-up steps. We report case-insensitive tokenized BLEU-4 scores with Moses\footnote{\url{https://github.com/moses-smt/mosesdecoder/blob/master/scripts/tokenizer/tokenizer.perl}}.

\paragraph{IWSLT14 German-English Translation} 
We use BPE to encode sentences with a shared vocabulary of 10K sub-word tokens. The phrase vocabulary of German is of size 0.4M and English of size 0.4M. We limit the maximum sentence length up to 256 words. We apply Adam with $\beta_1=0.9$ and $\gamma_2=0.98$ to train models for 100 epochs and select the best model parameters according to the model performance on the valid set. We use \textit{Transformer\_small} setting: embedding size as 512, FFN size as 1024, attention heads as 4, learning rate as 0.1, batch size as 6144, and dropout rate as 0.3. We use the warm-up strategy with 4000 warm-up steps.

\paragraph{WMT14 English-German Translation} 
We use BPE to encode sentences with a shared vocabulary of 10K sub-word tokens. The phrase vocabulary of German is of size 0.7M and English of size 0.4M. We limit the maximum sentence length up to 256 words. We apply Adam with $\beta_1=0.9$ and $\beta_2=0.98$ to train models for 50 epochs and select the best model parameters according to the model performance on the valid set. We use \textit{Transformer\_big} setting: embedding size as 1024, FFN size as 4096, attention heads as 16, learning rate as 0.1, batch size as 6144, and dropout rate as 0.1. We use the warm-up strategy with 4000 warm-up steps.

\section{Training Time Analysis}\label{time_space_analysis}

\begin{figure}[ht]
\begin{center}
    \includegraphics[width=0.9 \columnwidth]{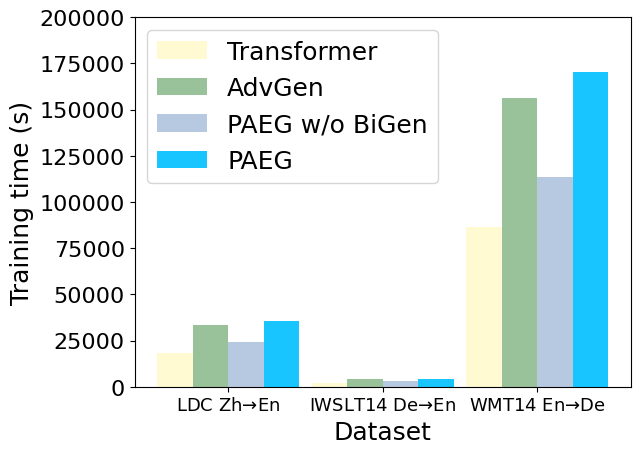}
    \caption{Training time of NMT models. All experiments on the three translation datasets are conducted on 8 NVIDIA 32G V100 GPUs and we set the batch size to fill the GPU memory. }
    \vspace{-10pt}
    \label{train_time}
\end{center}
\end{figure}

As our method uses augmented data, one concern is whether the training time increases too much. We record the time consumption of our method as well as AdvGen and Transformer. All experiments on the three translation datasets are conducted on 8 NVIDIA 32G V100 GPUs and we set the batch size to fill the GPU memory. 

The results are shown in Figure \ref{train_time}. The experiments show that AdvGen uses around the double time of training a Transformer, as it trains two (source and target) language models and generates adversarial data. Our method utilizes a pre-trained language model and thus saves the time of training the language model. Our method without bidirectional generation (BiGen) is faster than AdvGen. Even using bidirectional generation, our method is only slightly slower than AdvGen. Besides, the training time of \ourmethod{} is not exactly double of that of \ourmethod{} w/o BiGen, which is reasonable as the data from bidirectional generation do not deviate too much from the distribution of the original adversarial samples.

\end{CJK*}
\end{document}